\documentclass[runningheads]{llncs}
\usepackage{url}
\usepackage{hyperref}
\usepackage{setspace}
\usepackage{pgfplotstable}
\usepackage{float}
\usepackage{mathtools}
\usepackage{amsmath}
\usepackage{multirow}
\usepackage{adjustbox}
\usepackage{import}
\usepackage[shortlabels]{enumitem}
\usepackage{graphicx}
\usepackage{float}
\usepackage{lipsum} 
\usepackage{subcaption}
\usepackage{amsmath}
\usepackage{array}

\begin{document}
\title{LiveSchema: A Gateway Towards Learning on Knowledge Graph Schemas}


%
%
\author{Mattia Fumagalli\inst{1} \and
Marco Boffo\inst{2}\and 
Daqian Shi\inst{2} \and
Mayukh Bagchi\inst{2} \and\\
Fausto Giunchiglia\inst{2}
}

\authorrunning{F. Author et al.}
%
\institute{Conceptual and Cognitive Modelling Research Group (CORE)\\ Free University of Bozen-Bolzano, Bolzano, Italy \\
\email{mattia.fumagalli@unibz.it} \and
Department of Information Engineering (DISI)\\ University of Trento, Trento, Italy\\
\email{\{marco.boffo,daqian.shi,mayukh.bagchi,fausto.giunchiglia\}@unitn.it}
}
\maketitle              

\begin{abstract}
One of the major barriers to the training of \textit{algorithms} on \textit{knowledge graph schemas}, such as \textit{vocabularies} or \textit{ontologies}, is the difficulty that scientists have in finding the best input resource to address the target prediction tasks. In addition to this, a key challenge is to determine how to manipulate (and embed) these \textit{data}, which are often in the form of particular triples (i.e., \textit{subject}, \textit{predicate}, \textit{object}), to enable the learning process. In this paper, we describe the \textit{LiveSchema} initiative, namely a gateway that offers a family of services to easily access, analyze, transform and exploit knowledge graph schemas, with the main goal of facilitating the reuse of these resources in machine learning use cases. As an early implementation of the initiative, we also advance an online catalog, which relies on more than 800 resources, with the first set of example services.

\end{abstract}

\section{Introduction}
\label{intro}


Due to the emergence of a \textit{``Semantic Web''}, with the goal of enabling humans and computer systems to exchange and link data with unambiguous meaning, many repositories and catalogs have been created to host thousands of knowledge graph schemas \cite{kejriwal2019knowledge}, like \textit{vocabularies} and \textit{ontologies}. Thanks to these access points, more than tens of billions of facts spanning hundreds of linked datasets, covering different kinds of topics, are available on the web, and thousands of resources can be accessed, analyzed, and downloaded (see, for instance, the \textit{Datahub} and the \textit{LOV} \cite{vandenbussche2017linked} repositories\footnote{\url{https://datahub.io/}, \url{https://lov.linkeddata.es/dataset/lov/}}).

In the last years, the machine learning community has had an increasing interest in this kind of data. Knowledge graph schemas encode, indeed, a rich amount of information to be exploited in different settings, which span from data integration approaches \cite{dong2018data} to transfer learning solutions \cite{zhuang2020comprehensive}. For instance, the usage of ontology for supporting \textit{entity type recognition} and \textit{schema matching} has been largely explored in \cite{sleeman2015entity} and \cite{giunchiglia2020entity}. Similarly, works like \cite{kono2014transfer} and \cite{fumagalli2020ontology} have proved the high utility of vocabularies and ontologies in supporting \textit{zero shot learning for images recognition}. However, despite the enormous volume of available resources, scientists who are interested in reusing knowledge graph schemas for machine learning use cases, do not have a reference repository or a unique access point, where to gather them in the formats they need (e.g., tabular data created through \textit{propositionalization techniques} or \textit{embedding methods}) and to validate their utility given a target task (e.g., well-defined classes vs. classes with overlapping features). 

In this paper, we propose the \textit{LiveSchema} initiative, namely a gateway to be used as a reference access point of knowledge graph schemas, or graph-structured resources, like ontologies and vocabularies, for knowledge representation and machine learning use cases. We also advance a first implementation of this initiative as an online catalog that, besides gathering data accessible in reference state-of-the-art formats, like \texttt{RDF}\footnote{\url{https://www.w3.org/RDF/}} and \texttt{Turtle}\footnote{\url{https://www.w3.org/TeamSubmission/2011/SUBM-turtle-20110328/}}, provides visible indicators not previously harvested, and offers reference formats, e.g., the \textit{Formal Concept Analysis (FCA)} format \cite{ganter2012formal}, which are suitable for their application in statistical settings. The \textit{LiveSchema} implementation we propose here also leverages on the ideas presented in \cite{giunchiglia2020entity,giunchiglia2019knowledge,fumagalli2021ranking}, where an approach to analysing knowledge resources to address \textit{entity type recognition (ETR)} has been devised. The number of resources indexed by LiveSchema can constantly grow thanks to an aggregation function that keeps track of the updates coming from multiple source repositories and catalogs. The key purpose of LiveSchema is then to promote and facilitate the access, the analysis, and the reuse of knowledge graph schemas, not only in the context of data integration and knowledge representation but also in the training of statistical models, thus supporting scientists in the set-up of their machine learning algorithm.

The remainder of this paper is organized as follows: section 2 introduces how we formalize the notion of knowledge graph schemas in the context of LiveSchema and describes several evaluation metrics; then we discuss the data architecture and the data management of LiveSchema in section 3; section 4 presents a first implementation of the initiative; by means of a running example, we illustrate how a user can exploit the LiveSchema main functionalities in section 5; section 6 discusses implications and limitations of the proposed implementation; in section 7, we introduce the related work and then in section 8 we conclude our work.

\section{Knowledge Graph Schemas}

We call the resources collected by LiveSchema ``knowledge graph schemas'', namely data structures that, by taking inspiration from \textit{Formal Concept Analysis (FCA)}, can be formalized as: $K = \left \langle E_{K},P_{K},I_{K} \right \rangle$, with $E_{K} = \left \{ e_{1}, ..., e_{n} \right \}$ being the set of \textit{entity types} of \textit{K}, $P_{K} = \left \{ p_{1}, ..., p_{n} \right \}$ being the set of \textit{properties} of \textit{K}, and $I_{K}$ being a binary relation $I_{K} \subseteq E_{K} \times P_{K}$ that expresses which entity types \textit{are associated} to which properties. In the literature, knowledge graph schemas are commonly known as \textit{ontologies} or \textit{schemas \& vocabularies}, namely graph structured data that can be encoded as set of triples (e.g., \textit{subject}, \textit{predicate}, \textit{object}). We adopted the general notion of ``knowledge graph schema'' precisely to highlight our goal of accounting for all these different kinds of resources. LiveSchema will collect indeed data that span from vocabularies like DBpedia\footnote{\url{https://www.dbpedia.org/}}, through ontologies like SUMO\footnote{\url{https://www.adampease.org}}, to what are more informally known as schemas, like Schema.org\footnote{\url{https://schema.org/}}.

Given the above formalization,
we adopt a set of base metrics to describe any knowledge graph schema, which were previously introduced in \cite{giunchiglia2020entity} and are necessary to assess the \textit{informative value} of each given resource. These metrics are grounded on the core notion of \textit{Cue}, which was introduced for the first time in \cite{rosch1999principles}, and are 
applied to properties $p_{i}$, etypes $e_{i}$ and knowledge graph schemas $K_{i}$, and we use them to enable scientists in assessing each dataset before setting up any of the LiveSchema functionalities.

We define the \textit{cue validity of a property p to an entity type e}, also called $cue_{p}-validity$, as: 
\vspace{-0.5em}
\begin{equation}
\footnotesize
Cue_{p}(p, e) = \frac{PoE(p,e)}{|dom(p)|} \in [0, 1]
\end{equation}
where $|dom(p)|$ presents cardinality of entity types that are the domain of the specific property $p$; \textit{PoE(p, e)} is defined as:
\vspace{-0.5em}
\begin{equation}
\footnotesize
PoE(p,e) = \left\{\begin{matrix}1, if e \in dom(p)
\\ 
0, if e \notin dom(p)
\end{matrix}\right.
\vspace{-0.5em}
\end{equation}

$Cue_{p}(p, e)$ allows quantifying the centrality of a property in terms of how many times it is shared across the etypes of a graph.
$Cue_{p}(p, e)$ returns 0 if \textit{p} is not associated with \textit{e} and \textit{1/n}, where \textit{n} is the number of entities in the domain of \textit{p}, otherwise. In particular, if \textit{p} is associated to only one entity type its $cue_{p}-validity$ is maximum and equal to one.

Given the notion of $cue_{p}-validity$ we define the notion of cue validity of an entity type, also called $cue_{e}-validity$, as the sum of the cue validities of the properties associated with the entity, namely:
\vspace{-0.5em}
\begin{equation}
\footnotesize
Cue_{e}(e) = \sum_{i=1}^{|prop(e)|} Cue_{p}(p_{i}, e) \in [0, |prop(e)|]
\vspace{-0.5em}
\end{equation}

$Cue_{e}-validity$ allows determining the \textit{centrality} of an entity in a given K, by summing all its $Cue_{p}$. The value of $Cue_{e}-validity$ increases when the entity type maximizes the number of its properties with the members it categorizes. This metric is designed to assess how well an entity type is described by its properties. It can be applied when we need to determine the coverage potential of a given entity type (e.g., a definition of entity type \textit{person} with thirty properties is more likely reusable than a definition of \textit{person} with just two properties). 
Given the notion of $Cue_{e}-validity$, we capture the level of \textit{minimisation of the number of properties shared with other entity types, inside a K} with the notion of $cue_{er}-validity$, which we define as:
\vspace{-0.5em}
\begin{equation}
\footnotesize
Cue_{er}(e) = \frac{Cue_{e}(e)}{|prop(e)|} \in [0, 1]
\vspace{-0.5em}
\end{equation}

$cue_{er}-validity$ is critical for determining how well an entity type is defined, in terms of how many properties it shares with other entity types. If an entity type does not share properties with any other entity type, the $cue_{er}-validity$ is equal to 1. This metric is relevant to the prediction accuracy for a given entity type. For instance, if an entity type shares most of its properties with other entity types the probability of false-negative/positive increases.

The notions and terminology used for entity types, i.e., the notions of $Cue_{e}$ and $Cue_{er}$, can be straightforwardly generalized to K, generating the following metrics: 
\vspace{-0.5em}
\begin{equation}
\footnotesize
Cue_{k}(K)= \sum_{i=1}^{|E_{K}|} Cue_{e}(e_{i}) \in [0, |prop(K)|]
\vspace{-0.5em}
\end{equation}
The $Cue_{k}(K)$ is calculated as a summation of the cues of all the entity types of a given K, i.e., $E_{K}$, and returns the number of the properties of the K, i.e., $|prop(K)|$.
Following the formalization of $Cue_{er}$ we capture the level of \textit{minimisation of the number of properties shared across the entity types inside the schema} with the notion of $cue_{kr}-validity$, as:
\vspace{-0.5em}
\begin{equation}
\footnotesize
Cue_{kr}(K)= |prop(K)|/\sum_{i=1}^{|E_{K}|}prop(e_{i}) \in [0, 1]
\vspace{-0.5em}
\end{equation}
$Cue_{k}(K)$ and $cue_{kr}$ allows for a coarse-grained assessment of multiple reference Ks. A given K, can be then evaluated in terms of its number of entity types $E_{k}$, its number of properties $Cue_{k}(K)$ and its structural noise, in terms of $cue_{kr}$.

\section{Architecture}

\begin{figure}[t]
\vspace{-1em}
\centering
\includegraphics[width=0.7\linewidth]{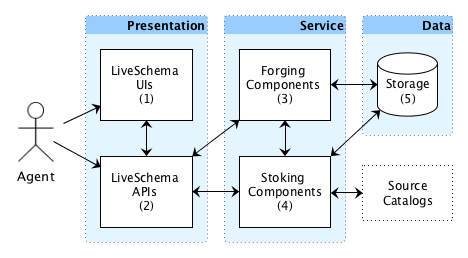}
\vspace{-1em}
\caption{Overview of the LiveSchema architecture.}
\vspace{-1.5em}
\label{archi}
\end{figure}

Fig. \ref{archi} provides a high-level representation of the LiveSchema architecture, which can be characterized by five main components. The (1) \textit{user interfaces UIs} and (2) the \textit{APIs} provide the main accesses to the gateway. The (3) \textit{stoking components} and the (4) \textit{forging components} cover the main novel contributions of the LiveSchema initiative, and offer the possibility to \textit{harvest}, \textit{generate} and \textit{process} data. The (5) \textit{storage} allows to collect in one place all the data collected by other catalogs, provided by the users or generated through the available services. We group 1, 2, 3, 4 and 5 in three main layers: the \textit{presentation layer}, the \textit{service layer} and the \textit{data layer}.

\vspace{-0.5em}
\paragraph{Presentation layer.} This layer enables a community of users: i) to maintain the whole gateway and its applications, and ii) to suggest and upload new resources or edit some already existing resources. The users involved in this layer will be expert knowledge engineers, software engineers, and data scientists that contribute to the development and evolution of the whole catalog. Moreover, a group of guest users will also be involved in the collaborative development of the storage, by uploading and editing some new datasets and, possibly, creating new input reference catalogs, following well-founded guidelines provided by the knowledge engineers that administrate the platform. The \textit{UIs} (1) allow users to access data and functionalities. Two types of user interfaces are present, namely front-end and back-end user interfaces. The former allows to access all the contents of the website through five main widgets: i) a menu with the top-level categories of the catalog, ii) a search form to easily access and browse datasets, iii) a showcase of the top services and a list of the source reference catalogs, and iv) recent activities. Differently, the back-end user interface will be accessible with credentials only and allows for the editing and submission of existing or new data, or it enables the usage of some more applications. The \textit{APIs} (2) allow the users to exploit the LiveSchema core functionalities by external code. Using the API, to provide some examples, developers will be able to: get formatted lists of vocabularies, with providers (namely, the agent who created the dataset), source catalogs, or other LiveSchema information; get a full representation of a dataset or other related information derived from the analysis of the dataset; search for datasets, providers, or other resources matching a query; create, update and delete datasets, with related metadata and information; get an activity stream of the recently changed dataset on LiveSchema, also obtaining the versioning information of each resource.

\vspace{-0.5em}
\paragraph{Service layer.} In this layer, the \textit{forging components} (3) represent a set of functionalities, which are aimed at the analysis and transformation of data, and the generation of new formats. The main goal here is to provide the users with accessible and easy-to-use services for the manipulation of knowledge graph schemas. Notice that the scope of one of the forging components is to transform each dataset in the reference formalization we introduced in section 2. Moreover, through another \textit{ad hoc} forging component it will be possible to apply the baseline metrics we discussed (i.e., $cue_e$ or $cue_k$). Similarly, the \textit{stoking components} (4) provide support for checking and gathering any new knowledge resource from a set of previously selected catalogs. Here manual and semi-automated processes for data insertion are supported. The former can be applied by any type of user, by submitting a new resource through a dedicated panel, but requires a review process from the administrator users. The latter is applied by selecting source catalogs as input and can be used to keep track of their updates. This stoking facility can be primarily customized by determining how many times the source catalogs must be checked and by defining what types of datasets can be collected and uploaded into the main storage. 

\vspace{-0.5em}
\paragraph{Data layer.} This layer is represented by the LiveSchema \textit{storage} (5). Here the goal is to offer an efficient method for collecting and classifying the data. Each imported dataset is connected with the related original publisher (if present) and is provided with information harvested from the source catalog. If the dataset was manually added by a LiveSchema user, the dataset description is provided in a submission phase. All the information about the dataset is then saved as \textit{metadata}. Notice that some data generated through the forging components can be very large. This allows for storing data only temporarily so that the user can download it if needed, and the server is not overloaded in long-term usage. 

\section{Implementation}

\subsection{System and Data}


The implementation of LiveSchema we propose here is grounded in the \textit{CKAN}\footnote{\url{https://docs.ckan.org/en/2.9/user-guide.html\#what-is-ckan}} open-source data management system which is widely recognized as one of the most reliable tools for managing open data. We concentrate on the fundamental \emph{distinction} in CKAN which informs the data architecture of LiveSchema, namely that between \emph{dataset} and \emph{resource}. A dataset is defined as a \emph{set of data} (e.g., BBC Sport Ontology) which may contain several resources representing the \emph{physical embodiment} of the dataset in different downloadable formats (e.g., BBC Sport Ontology in \texttt{TURTLE}, \texttt{FCA} formats). This distinction allows us, as a major advance from mainstream catalogs such as LOV \cite{vandenbussche2017linked}, to exploit \emph{fine-grained metadata} properties from the DCAT Application Profile for European Data Portals (DCAT-AP)\footnote{\url{https://www.w3.org/TR/vocab-dcat-2/}}, which makes a \emph{conceptually identical} distinction between \emph{dataset} and \emph{distribution}. The additional advantage of using DCAT-AP is that it organizes metadata into \emph{mandatory}, \emph{recommended} and \emph{optional} properties which is \emph{key} for facilitating different levels of semantic interoperability amongst data catalogs. 

We now elucidate the metadata specification, i.e. the selected metadata properties for datasets and distributions considered for the current version of LiveSchema:

\begin{itemize}
\vspace{-0.5em}
\item [i.] \textbf{Dataset}: 
\begin{itemize}
\item \texttt{MANDATORY}: \textit{description}, \textit{title};
\item \texttt{RECOMMENDED}: \textit{dataset distribution}, \textit{keyword}, \textit{publisher}, \textit{category};
\item \texttt{OPTIONAL}: \textit{other identifier}, \textit{version notes}, \textit{landing page}, \textit{access rights}, \textit{creator}, \textit{has version}, \textit{is version of}, \textit{identifier}, \textit{release date}, \textit{update}, \textit{language}, \textit{provenance}, \textit{documentation}, \textit{was generated by}, \textit{version}.
\end{itemize}
\item [ii.] \textbf{Distribution}: 
\begin{itemize}
\item \texttt{MANDATORY}: \textit{access url};
\item \texttt{RECOMMENDED}: \textit{description}, \textit{format}, \textit{license};
\item \texttt{OPTIONAL}: \textit{status}, \textit{access service}, \textit{byte size}, \textit{download url}, \textit{release date}, \textit{language}, \textit{update}, \textit{title}, \textit{documentation}.
\end{itemize}
\end{itemize}

\noindent Notice that the distinction between dataset and distribution metadata is \emph{non-trivial} in the sense that metadata properties like \textit{format}, \textit{license}, \textit{byte size} and \textit{download url} are associated to a distribution and \emph{not} to the dataset itself.

Our first observation concerns the two \emph{major} advantages which the aforementioned data distinction and metadata specification bring to LiveSchema. Firstly, metadata enforces \emph{`FAIR'ification} \cite{FAIR,FAIR4ML} of the KG schemas (which are `data' in this case), thus rendering them findable, accessible, interoperable, and reusable for the machine learning tasks which LiveSchema targets. Secondly, as a consequence of the first advantage, the metadata-enhanced KG schemas also play a pivotal role in initiating, enhancing, and sustaining \emph{reproducibility} \cite{RCR} which is \emph{key} for LiveSchema vis-à-vis the target machine learning ecosystem in which it participates.

Our second observation concerns the future extensibility of the metadata specification of LiveSchema. The starting distinction between dataset and distribution can help bootstrap the extension of the initial metadata specification to \emph{ontology-specific} metadata which, \emph{mutadis mutandis}, preserves the same distinction via the notions of \emph{ontology conceptualization} and \emph{ontology serialization} \cite{OMV}. One of the \emph{key} advantages of using ontology-specific metadata in LiveSchema is that the user can perform a highly customized (conjunctive) search, for instance, even at the level of logical formalism or ontology design language, thus retrieving the most compatible schema for the machine learning task at hand. In this direction, we plan to exploit the OMV \cite{OMV} and the MOD \cite{MOD} ontology metadata proposals in the immediate future.

\subsection{Data Collection and Development}

At the current state LiveSchema relies on four main state-of-the-art catalogs, namely LOV, DERI\footnote{\url{http://vocab.deri.ie/}}, FINTO\footnote{\url{http://finto.fi/en/}} and a custom catalog\footnote{\url{http://liveschema.eu/organization/knowdive}}, where some selected resources are stored. 


Each catalog is associated with a provider, which is the person or the organization that uploaded the dataset and is in charge of its maintenance in the source catalog. From each catalog, multiple datasets have been individually scraped and uploaded in an automated way. Currently, the catalog providing most of the datasets is LOV, being one of the most widely used catalogs of vocabularies in the semantic web context. 

Among the scraped dataset a check is performed to ensure that LiveSchema is not breaking any license agreement. Currently, five kinds of licenses are admitted given their restrictions (all of them are part of the \textit{Creative Commons}\footnote{\url{https://creativecommons.org/}} initiative). These license constraints need to be checked since we both provide access and we manipulate their content to provide the following resources. 
As we parse them from their source from various sets of formats, we serialize them into the most common ones, namely \texttt{RDF} and \texttt{Turtle}. More advanced output formats are the five ones generated through the processing operations enabled by the LiveSchema services, namely \texttt{CSV} (where all the triples and metadata of the input knowledge graph schema are stored in a datasheet format), \texttt{CUE} (where all the cue metrics are provided), \texttt{FCA} (i.e., the FCA transformation matrix result), \texttt{VIS} (the format that can be used to enable visualization services functionalities), \texttt{EMB} (the format used to generate a statistical model based on a knowledge embedding process). 

Besides allowing users to manually upload datasets through the admin functionalities, LiveSchema supports an automated evolution process, which mainly consists of a parsing phase. Few check-points are released for administrators to supervise the output of the automatic processes. The parsing process is very simple, it is executed iteratively and has the goal of producing two main outputs, namely a set of serialized datasets and a set of parsed datasets. The first output is produced by scanning the datasets list and parsing it using RDFlib python library\footnote{\url{https://github.com/RDFLib/rdflib}}, namely a library that is used to process RDF resources. Here the produced output is used to generate more standard reference formats, which, in the current setting are represented by \texttt{RDF} and \texttt{Turtle}. We also allow for the generation of an excel file encoding all the information (e.g., triple and metadata) about the dataset to easily enable the other applications provided by the catalog. In this step, the key role of the admin in charge of the parsing process is to edit the dataset list in order to filter out undesired datasets and parse only the ones that are required. The second output is then produced by scanning each triple of the input dataset. The filtering among the predicates to specify the focus of the dataset is applied just before the application of some services.

All the datasets that are gathered from the source catalogs and uploaded to LiveSchema can be then transformed and used as input for the available functionalities. In the current LiveSchema setting there exist six main functionalities, which are 1) \textit{FCA generator}, namely the process by which data can be converted in the FCA format (\texttt{FCA}); 2) \textit{CUEs generator}, i.e., the process by which the CUEs (as defined in Section 2) are generated and encoded in the \texttt{CUE} format; 3) \textit{Visualization generator}, namely the process by which the input data can visualized and analysed (see \texttt{VIS} format; 4) \textit{Knowledge Embedder}, i.e., the application by which a model can be created out of the input data, by applying one or some of the libraries provided by the PyKEEN package\cite{ali2020pykeen} (see \texttt{EMB} reference format); 5) the \textit{Query Catalog} service, which allows to run SPARQL queries\footnote{\url{https://rdflib.readthedocs.io/en/stable/intro\_to\_sparql.html}}; 6) the \textit{knowledge graph visualizer}, namely an implementation of the WebVOWL\footnote{\url{http://vowl.visualdataweb.org/webvowl.html}} library. This set of functionalities can be easily accessed and reused by means of APIs services, and can also be easily extended, e.g., 4), 5) and 6) can be run directly using \texttt{.rdf} files as input. Each functionality may require an \textit{ad hoc format} to produce the output, and, in some cases, it may have some dependencies with the input format of other functionalities, e.g., 1), 2) and 3) involves new formats.

\section{Usage Example}

The scope of this section is to show how the LiveSchema processing component works. By means of a running example, we illustrate how a user can exploit main functionalities. All the described operations can be directly tested by exploring and using the LiveSchema ecosystem, which is accessible at \url{http://liveschema.eu/}. The admin functionalities can be accessed and tested at \url{http://liveschema.eu/user/login}, by using \texttt{`reviewer'} as \textit{admin/password}.

\subsection{Analyzing Knowledge Graph Schemas}

As an example scenario, suppose we need to run a standard \textit{entity type recognition task}, as it is described in \cite{giunchiglia2020entity} and \cite{sleeman2015entity}, where we may need to recognize objects of the type `Person' across a set of multiple tabular data, coming, for instance, from an open data repository. This may involve that we need to find a reference knowledge graph schema with i) the target class and corresponding label; ii) possibly a huge number of properties for the target class, in order to increase the probability to match some of the input test properties; iii) possibly a low number of overlapping properties, in order to decrease the number of false-negative/positive.

A LiveSchema user can perform a simple search across the available datasets that are present in the catalog and then run an analysis to select the best. The LiveSchema search facility exploits the CKAN search engine that allows for a quick ‘Google-style’ keyword search. All the datasets, providers, and group fields are searchable and the users can use all of them to research the desired entity. Thanks to this search functionality it is possible to provide a complete and customizable service to the scientist looking for the desired ontology. The basic supported search options are i) search over all the datasets attributes, namely by using any of the applied metadata; ii) full-text searching; iii) fuzzy-matching, namely an option to search for closely matching terms instead of exact matches; iv) search via API.

Now, suppose that the user identifies three candidate resources for the goal ETR task, namely \textit{Schema.org}\footnote{\url{https://schema.org/}} (reference standard to support the indexing of web documents by \textit{Google}), \textit{FOAF}\footnote{\url{http://xmlns.com/foaf/spec/}} (a widely used vocabulary in the context of social networks) and the \textit{BBC sport ontology}\footnote{\url{https://www.bbc.co.uk/ontologies/sport}} (the ontology used by the BBC to model supports events, and roles). The next step is to access every single dataset and check its meta-information, which can be done by firstly generating the \texttt{FCA} format for the selected resources.


Each LiveSchema dataset has a dedicated page collecting its information, and where each processing functionality can be accessed. Here information about the related source catalog is provided as well, and the available standard reference formats can be downloaded. The FCA functionality can be accessed through the corresponding tab and allows for the generation of the corresponding matrix for each given input knowledge graph schema. On the FCA service page is also possible to customize the generation of the matrix by filtering the target predicates. Then, multiple insights can be extracted by using the functionalities represented by each tab on the dataset page. By downloading all the cue information comparison between the three representations of `Person', provided by each ontology can be run. Table 1 represents the cue values for the given resources. A brief benchmark shows that in Schema.org, even if the cue of Person is not at the top, the given class has a high centrality with a score of 23. 

\vspace{-1em}
\begin{figure}[ht]
\centering
\begin{subfigure}[b]{0.4\linewidth}
\small
\centering
\begin{tabular}{ccc}
\hline
\textit{Class} & \textit{$Cue_{e}$} & \textit{$Cue_{er}$} \\ \hline
`Person' - Schema.org & \textbf{23} & \textbf{0.81} \\
`Person' - FOAF & \textbf{3} & \textbf{0.82}\\
`Person' - BBC & \textbf{0.73} & \textbf{0.75} \\ \hline
\end{tabular}
\vspace{1em}
\end{subfigure}
\hspace{5mm}
\begin{subfigure}[b]{0.4\linewidth}
\includegraphics[width=1\linewidth]{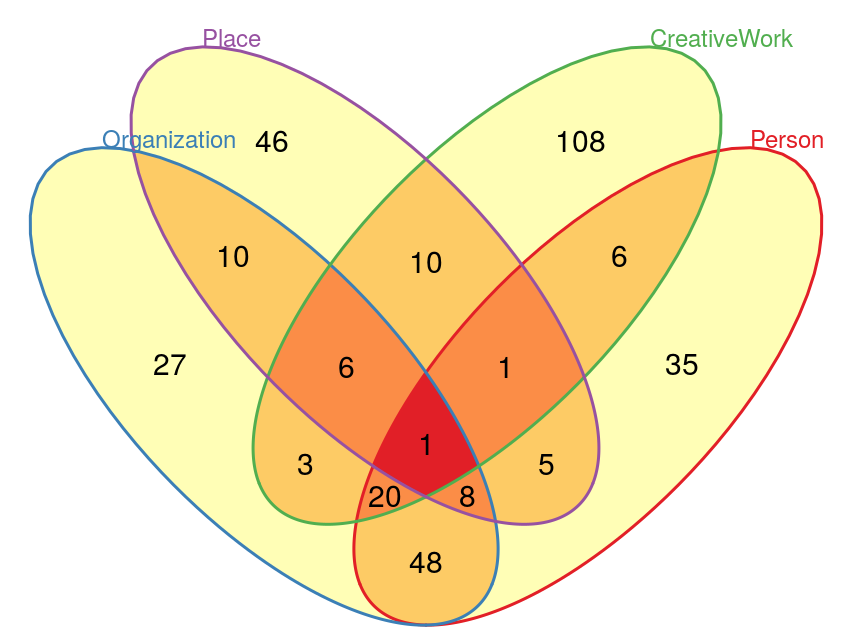}
\caption{}
\end{subfigure}
\caption{(a) Cue values for the class `Person'; (b) Etypes properties intersection visualization (Organization, Place, CreativeWork and Person).}
\label{double}
\vspace{-1em}
\end{figure}

Besides the quantification of the cues, further analysis can be run by visualizing the intersection of some of the top classes of the given resources. Figure \ref{double}.b represents an example of \textit{knowledge lotus} that can be extracted by the input resources. Knowledge lotuses are \textit{venn diagrams} that can be used to focus on specific parts of the input resources and they are particularly useful to represent the diversity of etypes in terms of their (un-)shared properties. 
The yellow petals of the lotus show the number of properties that are distinctive for the given etype. In the example, Person has 35 un-shared properties. The different shades of orange represent the number of properties shared with other etypes (for instance, there is 1 property that is shared with all the etypes). 

\begin{figure}[ht]
\centering
\includegraphics[width=1\linewidth]{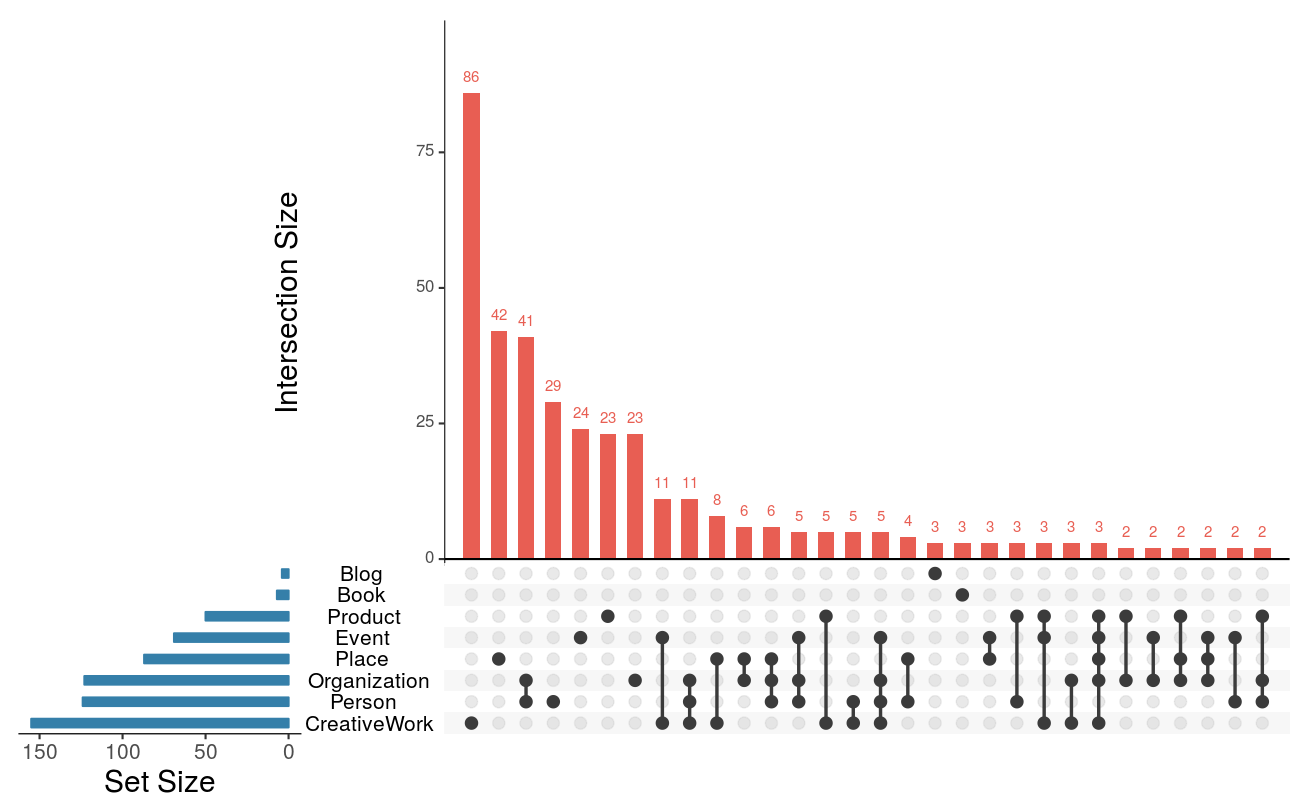}
\caption{Etypes properties intersection: the UpSet visualization.}
\label{09}
\vspace{-1em}
\end{figure}

Further analysis can be run by applying the UpSet (multiple set) visualization facilities, which allows us to analyze the intersections between etypes, by selecting more than 6 sets (the limit for knowledge lotuses). LiveSchema allows for both knowledge lotuses and UpSet visualization by embedding the functionalities of the intervene visualization environment\footnote{\url{https://intervene.shinyapps.io/intervene/}}. This environment was created for the visualization of multiple genomic regions and gene sets (or lists of items).
The main goal of the provided visualization options is to facilitate the analysis and interpretations of the input resource. An illustrative example of the representation of a resource by means of the UpSet module is provided by Figure \ref{09}. Here 8 etypes are selected. The blue bars on the left show the size of the etypes in terms of the number of properties. The black dots identify the intersections between the etypes and the red bars on top f the chart show the size of the properties intersection set. 

\subsection{Embedding Knowledge Graph Schemas}

Once the scientist has selected her resource, she is ready to embed it and generate a statistical model out of it. Notice that, in the current release of LiveSchema we allow for distributional embedding techniques only. In this current setting LiveSchema relies on a recent library collecting most of the state-of-the-art techniques for graph-embedding, namely the PyKEEN library \cite{ali2020pykeen}. PyKEEN is a widely used solution for generating custom embedding models. It allows for selection across a wide range of training approaches with multiple parameters and will output a \texttt{.pkl} file which can be directly imported inside ML pipelines.

Notice that in LiveSchema we have datasets encoding knowledge graph schemas with no instance data (e.g., we have the DBpedia schema, but we do not have the so-called ABOX). This did not prevent us to adapt the embedding process and focus on the schema level only (relying on relational data we always have, indeed, triples: \textit{heads}, \textit{tails} and \textit{relations}). This, besides opening the possibility to test a new application scenario, does not exclude the possibility to apply the standard approach where populated schemas are used as input. 

\section{Discussion}

We believe that LiveSchema is a novel initiative that supports researchers to study knowledge graph schemas for both knowledge representation tasks (e.g., designing a knowledge base for enabling the interoperability between systems), and machine learning tasks, in particular the ones that rely on relational data structures for training their models. In this section, we discuss the implications of this novel initiative. Moreover, by identifying the limitations of our current setting, we also discuss opportunities for future work.

\subsection{Implications}
Firstly, our proposed LiveSchema supports scientists to find the knowledge graph schemas they need. By leveraging the updates of some of the best state-of-the-art catalogs, LiveSchema offers an aggregation service that allows keeping track of the evolution of the knowledge representation development community in one place. Notice that this does not aim to substitute the function of each single source catalog. The scientist can indeed access the source catalog and related \textit{ad hoc} services, if needed, directly from the LiveSchema gateway, this being also an opportunity of increasing the visibility of the vocabularies themselves. 

Another key point is that LiveSchema offers an opportunity to bridge the gap between two key artificial intelligence communities, namely the knowledge representation and the machine learning community. While most of the data that are present in LiveSchema are indeed in a format that is compliant with the knowledge representation applications requirements, each dataset can be also transformed so that it can be easily employed in machine learning set-ups. The analysis and embedding facilities offer further support in this direction. We believe that this is a way of supporting the exploitation of the huge amount of work done by the community and of making the knowledge graph schema more accessible to machine learning scientists. 

Moreover, we implement state-of-the-art libraries to support data scientists in the data analysis and preparation phases. Most of the implemented libraries require coding and development skills, which will limit the usage of data scientists. To solve this issue, LiveSchema offers a platform that unites data analysis, data preprocessing, and machine learning model deployment and makes them easily accessible and usable.

Finally, the overall project was also devised to pave the way for large case studies. The integration of knowledge representation and machine learning scenarios may imply, for instance, a different way of designing knowledge graph schemas, with a different focus on some of their features or constraints (e.g., the number of properties to be used for describing a class or the overlapping between etypes). As another example, scientists may rank the knowledge resources according to their utility for a target task.

\subsection{Limitations}

Developing LiveSchema as a community of data scientists that exchange and reuse data to the benefit of the AI community is our long-term objective and this triggers the agenda for immediate future work. In order to achieve this goal, with the current set-up, there is still a gap that needs to be bridged.

As long as the LiveSchema access point will grow, serious issues about the scalability of the approach will need to be handled. For instance, in the current implementation is possible to keep track of the dataset versioning provided by each source catalog, but the automatic checking of duplicate resources coming from different catalogs is missing. 

Another pending issue, which is part of the agenda for the immediate future work, is the definition of a processing component functionality that enables users to work with multiple datasets together, this being a key option, especially for data integration tasks and the evolution of more robust machine learning models. A possible way of implementing this functionality will be to develop a new version of the current FCA conversion process, where multiple datasets can be given as input and then merged, by computing their similarities, into one single file. The output file will be then used as a single resource containing the information of all its component datasets.

\section{Related Work}

The presented work builds primarily upon the large amount of work in recent years on the generation of repositories, for maintaining and reusing knowledge resources, like in particular, ontologies, or more generally, vocabularies. Most of these resources have been created with the purpose of supporting knowledge engineers in the selection, re-usage, and sharing of such resources \cite{issues,ontopublish}. Based on an in-depth analysis of some of the most used catalogs, which were also aimed at selecting the list of candidate catalogs by which populating and evolving LiveSchema, we identified some initiatives that are more related to our approach, according to the functionalities they provide, the metadata, the way of extending the catalog, the number of datasets, the way of browsing and searching the resources and the available formats.

At the top of the list, we applied \textit{LOV (Linked Open Vocabulary)}. LOV, started in 2011, now hosted by the \textit{Open Knowledge Foundation}\footnote{\url{https://okfn.org/}}, is a high-quality catalog of reusable vocabularies for the description of data on the web. It provides a choice of several hundreds of vocabularies, based on quality requirements including URI stability and availability on the web. It relies on standard formats and publication best practices, quality metadata, and documentation. As a distinctive feature, LOV makes visible indicators that are not provided by other catalogs, such as the interconnections between vocabularies, and the versioning history along with past and current editors (individual or organization). Other relevant projects include: \textit{Finto (Finnish thesaurus and ontology service)}; \textit{BARTOC Skosmos\footnote{\url{http://bartoc-skosmos.unibas.ch/en/}}}, namely \textit{SKOS Vocabulary}; \textit{DERI} project and \textit{ANDS-RVS(Australian National Data Service-Research Vocabularies Australia)\footnote{\url{https://vocabs.ands.org.au/}}}.

Differently from the listed projects, LiveSchema presents features, such as the capability to keep evolving its storage by leveraging on the updates of these state-of-the-art catalogs, and the processing component services, which are aimed at keeping track the evolution of existing catalogs and reusing knowledge graph schemas in the context of machine learning use cases. Our work, then, also exploits many research results provided by the more general area of \textit{relational data processing and transformation} research. 

\section{Conclusion}



In this paper, we proposed the \textit{LiveSchema} initiative and its first early implementation, which is accessible at \url{http://liveschema.eu/}. This contribution mainly rises from the idea of leveraging the gold mine of data collected and produced by the KR community in the last years for addressing the increasing need for high-quality semantic data from the ML community.

By implementing a continuously evolving aggregation facility, which keeps track of the updates from selected existing catalogs, LiveSchema offers a family of services to easily access, analyze, transform, and re-use knowledge graphs schemas. The final goal is to advance the state of the art in the creation of key reference access points for analyzing and reuse knowledge graph schemas, not only in a classical knowledge representation context, e.g., to enable reasoning and/or interoperability but also for machine learning use cases.

\vspace{-0.5em}


\vspace{1.5em}

%
%
%
%

\bibliographystyle{unsrt}
\bibliography{MAIN}

\end{document}